\documentclass[a4paper,fleqn]{cas-dc}

\usepackage{amsmath,amssymb,amsfonts}
\usepackage{algorithmic}
\usepackage{algorithm}
\usepackage{array}
\usepackage{bbding}
\usepackage{booktabs}
\usepackage[caption=false,font=normalsize,labelfont=sf,textfont=sf]{subfig}
\usepackage{textcomp}
\usepackage{color}

\usepackage{stfloats}
\usepackage{url}
\usepackage{bm}
\usepackage{verbatim}
\usepackage{graphicx}
\usepackage{cite}
\usepackage{multirow}
\usepackage{dsfont}
\usepackage{xpatch}
\usepackage[numbers]{natbib}

\def\tsc#1{\csdef{#1}{\textsc{\lowercase{#1}}\xspace}}
\tsc{WGM}
\tsc{QE}
\tsc{EP}
\tsc{PMS}
\tsc{BEC}
\tsc{DE}

\begin{document}
\let\WriteBookmarks\relax
\def\floatpagepagefraction{1}
\def\textpagefraction{.001}
\shortauthors{Liu et~al.}

\title [mode = title]{RCCFormer: A Robust Crowd Counting Network Based on Transformer}                      

\tnotetext[1]{This work was supported in part by the National Natural Science Foundation of China under Grant 62271418, and in part by the Natural Science Foundation of Sichuan Province under Grant 2023NSFSC0030 and 2025ZNSFSC1154,  and in part by the Postdoctoral Fellowship Program and China Postdoctoral Science Foundation under Grant Number BX20240291.}


\author[1]{Peng Liu}[style=chinese,
                        orcid=0009-0005-2577-7371]

\credit{Conceptualization of this study, Methodology, Software}

\affiliation[1]{organization={School of Information Science and Technology},
                addressline={Southwest Jiaotong University}, 
                city={Chengdu},
                citysep={}, 
                postcode={611756}, 
                country={China}}

\author[1]{Heng-Chao Li}[style=chinese]
\cormark[1]
\author[1]{Sen Lei}[style=chinese]
\author[1]{Nanqing Liu}[style=chinese]
\author[2]{Bin Feng}[style=chinese]
\author[3]{Xiao Wu}[style=chinese]


\credit{Data curation, Writing - Original draft preparation}

\affiliation[2]{organization={School of Physical Education},
                addressline={Southwest Jiaotong University}, 
                city={Chengdu},
                citysep={},
                postcode={611756}, 
                country={China}}

\affiliation[3]{organization={School of Computing and Artificial Intelligence},
                addressline={Southwest Jiaotong University}, 
                city={Chengdu},
                citysep={},
                postcode={611756}, 
                country={China}}


\cortext[cor1]{Corresponding author}


\begin{abstract}
Crowd counting, which is a key computer vision task, has emerged as a fundamental technology in crowd analysis and public safety management. However, challenges such as scale variations and complex backgrounds significantly impact the accuracy of crowd counting. To mitigate these issues, this paper proposes a robust Transformer-based crowd counting network, termed RCCFormer, specifically designed for background suppression and scale awareness. 
The proposed method incorporates a Multi-level Feature Fusion Module (MFFM), which meticulously integrates features extracted at diverse stages of the backbone architecture. It establishes a strong baseline capable of capturing intricate and comprehensive feature representations, surpassing traditional baselines.
Furthermore, the introduced Detail-Embedded Attention Block (DEAB) captures contextual information and local details through global self-attention and local attention along with a learnable manner for efficient fusion. This enhances the model's ability to focus on foreground regions while effectively mitigating background noise interference.
Additionally, we develop an Adaptive Scale-Aware Module (ASAM), with our novel Input-dependent Deformable Convolution (IDConv) as its fundamental building block. This module dynamically adapts to changes in head target shapes and scales, significantly improving the network’s capability to accommodate large-scale variations. 
The effectiveness of the proposed method is validated on the ShanghaiTech Part\_A and Part\_B, NWPU-Crowd, and QNRF datasets. The results demonstrate that our RCCFormer achieves excellent performance across all four datasets, showcasing state-of-the-art outcomes. The code will be available at \url{https://github.com/lp-094/RCCFormer}.






\date{}
\maketitle

\section{Introduction}

Crowd counting is a key and challenging task that employs vision-based methods to estimate the number of people in unconstrained scenarios. In recent years, due to rapid population growth, crowd counting has gained significant attention for its wide-ranging applications in public crowd monitoring\cite{LUO201627}, video surveillance\cite{XIA2016672} and traffic control\cite{meng2023caal}. With the development of vision methods, a considerable amount of research has been conducted \cite{li2018dilated,liu2019context,lin2022boosting}.

\begin{figure}[!h]
	\centering
	\includegraphics[width=3.3in]{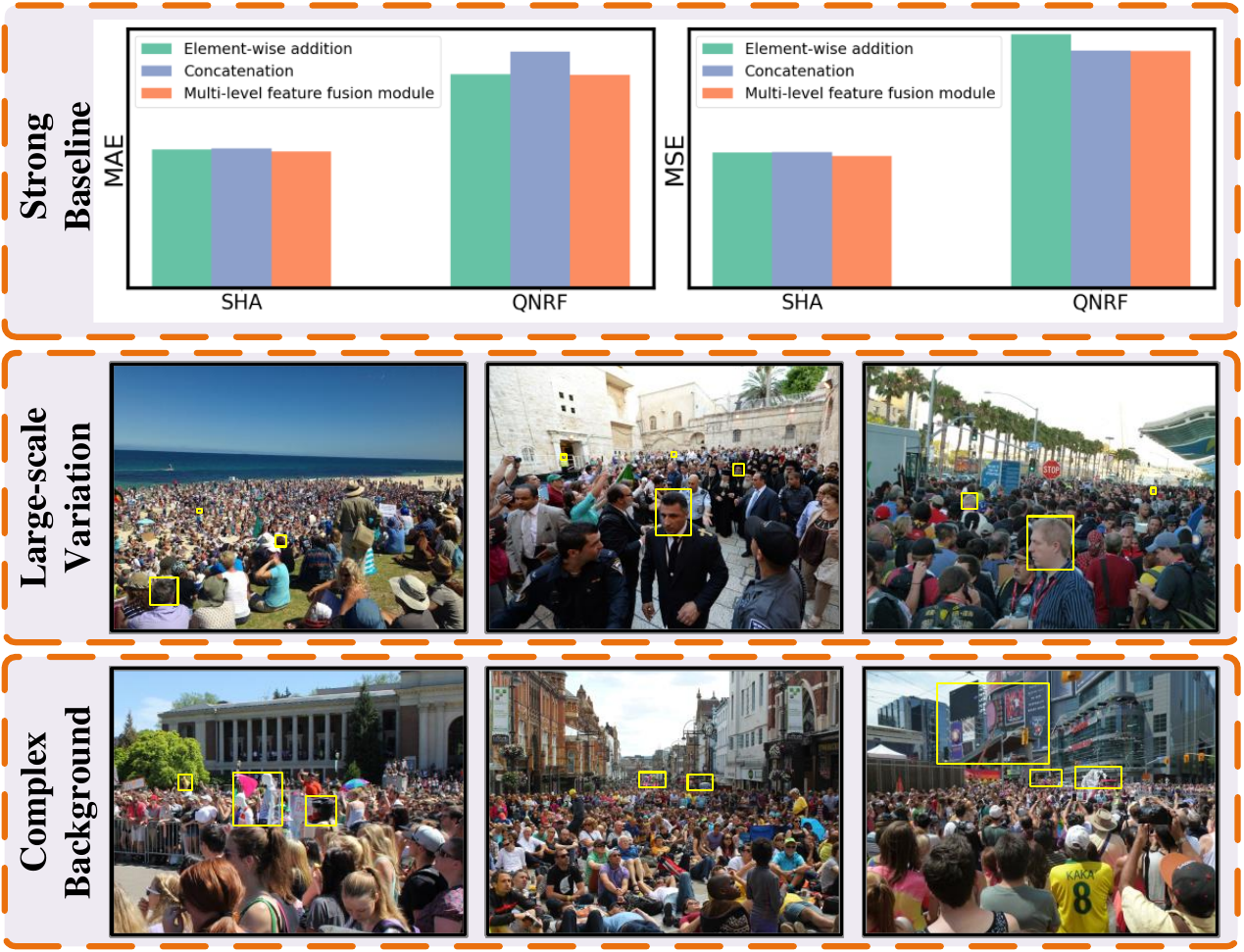}
	\caption{The effectiveness of strong baseline and typical crowd counting challenges: (1) \textbf{Strong Baseline:} (top) strong baseline significantly improves several datasets compared to the vanilla method. (2) \textbf{Large-scale Variation:} different heads pose substantial scale variations, with the largest being several times larger than the smallest (middle); (3) \textbf{Complex background:} billboards, cluttered backgrounds, and other distractions complicate the task of counting (bottom). }
	\label{fig_1}
\end{figure}

Among these methods, transformer-based methods \cite{lin2022boosting,tian2021cctrans} tend to perform better than CNN-based methods. However, the transformer backbone \cite{dosovitskiy2020image,wang2021pyramid} is still limited by extracting single-level features. Relying solely on high-level features can cause the network to lose many detailed characteristics such as accurate location information of crowds. To solve this issue, CCTrans \cite{tian2021cctrans} employs element-wise addition for multi-layer feature fusion, and PTCNet\cite{liu2023multi} utilizes a cascading approach. However, both methods overlook the inherent information discrepancy in these approaches. To this end, we construct a simple but effective method to utilize both high-level and low-level information fully. We use the features from the last three stages of the backbone, process them through a series of operations, and input them into the cross-attention for selective fusion. We refer to this approach as a Strong Baseline. From Fig. \ref{fig_1} (first row), it is evident that the strong baseline significantly improves several datasets compared to the vanilla method. For specific details, please refer to the Sec. \ref{sec:strong baseline}.

After obtaining a strong baseline, we can extract sufficiently robust features for subsequent processing. However, due to the counting task being affected by the imaging distance of the camera and complex backgrounds, the following issues still exist: 

\textbf{(1) Large scale variation.} Scale variation is one of the main challenges in crowd counting. As shown in Fig. 1 (second row), it can be observed that heads closer to the camera display larger scales, while those farther away seem smaller. This notable scale disparity among different head instances presents a challenge in thoroughly exploring all scales. To mitigate this challenge, numerous methods involve designing CNN architectures that integrate multi-scale feature representations by fusing features from different layers\cite{liu2023multi}, branches\cite{bai2020adaptive}, or multi-columns\cite{liu2022lwcount,yang2020counting}. However, traditional CNNs have inherent drawbacks, such as fixed-size and limited receptive fields, which restrict their ability to explore large scales and weaken their capability to perceive shape variations. This leads to poor performance in scenarios with large-scale variation and a variety of head target alterations. Hence, modeling adaptive and large-scale perception becomes essential for adapting the rapid intra- and inter-image scale and shape variety.

\textbf{(2) Confusion between foreground and background.} Complex background also poses a significant challenge to accurate counting. As depicted in Fig.1 (third row), objects such as mannequins on billboards, and complex-colored garments may be mistakenly identified as heads. Many existing methods leverage CNN-based attention mechanisms, such as channel attention, spatial attention, and CBAM, to mitigate this issue. For instance, inspired by spatial attention mechanism, Sindagi et al.\cite{sindagi2019ha} developed a global attention module to focus on the channel dimension of feature maps, as well as a spatial attention module to suppress irrelevant features. Similarly, Wang et al. \cite{wang2020sclnet} proposed spatial attention branch and channel attention branch, effectively integrating them through dot-product operations. However, these methods still struggle to yield satisfactory results due to the independent weighting of channel/spatial locations and the lack of effective foreground object modeling. 
Transformers can mitigate this issue to some extent, as they can model global contextual dependencies, providing global information to explore the correlations between the foreground and background. However, they are limited in modeling local context, which contains the fine-grained features necessary for distinguishing between heads and complex backgrounds. Therefore, the effective integration of global and local features is particularly important.

Based on the preceding analysis, we further propose a robust crowd counting network (RCCFormer) based on the strong baseline. To mitigate the \textbf{confusion between foreground and background}, we propose a Detailed Embedding Attention Block to better extract human heads from complex backgrounds. Specifically, we leverage self-attention to extract global contextual information and employ our local attention to mine local detailed information, followed by selective fusion in a learnable manner. Our local attention efficiently captures detailed information at minimal computational cost compared to self-attention. Moreover, unlike CNNs, it generates dynamic weights, enhancing adaptability and effectively bridging the semantic gap with Transformers \cite{han2021connection}. In order to solve the \textbf{large-scale variation} issue, we introduce Input-dependent Deformable Convolution (IDConv) in our framework. IDConv adaptively learns deformation variables and weights for convolutions through feature learning. Compared to vanilla convolutions and deformable convolutions, IDConv better models scale and shape-varied head targets. Subsequently, we construct an Adaptive Scale-Aware Module using IDConvs with varying dilation rates to enhance the learning of large-scale feature representations.

In summary, the contributions of this paper are four-fold:

\begin{itemize}
	\item[$\bullet$] We establish a strong baseline that uses a multi-level feature fusion module to integrate the features of multiple stages in the backbone. It achieves significantly better performance than the vanilla baseline.
	\item[$\bullet$] We introduce the detail-embedded attention block (DEAB), which incorporates local attention with dynamic attributes to bridge the semantic gap with global self-attention. Through a learnable manner, we seamlessly blend local detail information with global context information to adaptively suppress background noise.
	\item[$\bullet$] We construct an adaptive scale-aware module (ASAM) based on our proprietary Input-dependent Deformable Conv (IDConv), which can learn convolutional weight adjustments and shape variations simultaneously, leading to more effective modeling of geometric transformations. This module enables more effective capturing of multi-scale feature representations.
	\item[$\bullet$] Extensive experiments across four popular benchmarks including ShanghaiTech Part\_A and Part\_B, UCF-QNRF, and NWPU-Crowd, show that our proposed RCCFormer reaches new state-of-the-art results.
\end{itemize} 

The remainder of this paper is organized as follows. Section~\ref{sec:related_work} reviews some important works related to our RCCFormer. In Section~\ref{sec:method}, we describe the implementation details of our RCCFormer. Section~\ref{sec:experiments} contains extensive experiments on various crowd counting datasets. In Section~\ref{sec:conclusion}, we summarize the whole paper.

\section{Related Work} \label{sec:related_work}

\subsection{Crowd Counting}
Existing deep learning-based crowd counting approaches are mainly divided into three categories: detection, regression counting, and density estimation. The detection-based methods \cite{liu2019point,wang2021self} involve detecting bounding boxes around human heads in the image and calculating the number of boxes to obtain the counting result. However, its performance is seriously affected in occlusion and dense scenes. The regression-based methods \cite{chen2013cumulative,wang2015deep} do not need annotation information and directly obtain the counts. While these methods may be straightforward, their precision in counting leaves something to be desired. Density estimation is currently the most mainstream method \cite{li2018dilated,gao2020pccnet}, where a density map is generated through the network, and the sum of density values is the final counting result. More recently, researchers have developed various techniques and models around density estimation to improve crowd counting performance, mainly including multi-scale models and attention models -- whose details are described below:

{\bf{Multi-scale Models:}} To enhance the network's adaptation to scale variation, early research adopted a multi-column method \cite{liu2022lwcount,yang2020counting} that effectively merge multi-scale features through branches with different receptive fields. Some researchers designed pyramid structures to explore multi-scale features. Sindagi et al. \cite{sindagi2017generating} proposed a Context Pyramid Convolutional Neural Network (CP-CNN) to enhance counting performance by constructing global and local multi-scale contexts. SASNet \cite{song2021choose} utilized a feature pyramid network to learn cross-scale and feature-level correlations. The final prediction is the weighted average of individual predictions from different levels. Additionally, using dilated convolutions with different dilation rates to construct multi-scale feature representations is also a commonly used method. STNet \cite{wang2022stnet} leveraged dilated convolutions to construct a tree structure, enabling the hierarchical parsing of coarse-to-fine crowd regions. However, they lack the ability to adaptively perceive different scales.

{\bf{Attention Models:}} The attention mechanism plays an important role in crowd counting research by allowing the network to focus on the most relevant information within the input. Designing an appropriate attention module can effectively improve the performance of crowd counting. \cite{liu2019adcrowdnet} proposed an attention-injective deformable convolutional network, which incorporates an attention map generator to highlight foreground areas. CAFNet \cite{wang2023context} designed a guidance attention fusion to compensate spatial information of low-level and high-level feature maps. Guo et al. \cite{guo2023object} proposed a group channel attention and learnable graph attention to suppress the background noise.

Despite the significant progress that has been achieved with the aforementioned CNN-based multi-scale and attention methods, the limited receptive field of CNNs constrains their ability to capture global context in large-scale and complex scenes, impacting the accuracy of crowd counting. Transformer-based networks can mitigate this issue to some extent, as they have the ability to model global contextual dependencies. Therefore, this paper primarily utilizes Transformers as the basic architecture.

\subsection{Vision Transformer}
Since Dosovitskiy et al. \cite{dosovitskiy2020image} first introduced the Transformer into the field of computer vision, a large number of Transformer-based methods have been developed for downstream tasks \cite{carion2020end,9654169,tian2024hirenet,lei2024exploring,LV2024109542,NING2025109768}. Drawing inspiration from the feature pyramid architecture, Wang et al. \cite{wang2021pyramid} designed a multi-scale hierarchical Transformer structure for dense prediction tasks. Subsequently, they further developed PVTv2 \cite{wang2022pvt} by employing overlapping patches and convolutional feedforward networks to handle images of arbitrary scales. The DETR series \cite{carion2020end,zhu2020deformable,cheng2024adin} used a CNN backbone for visual feature extraction followed by Transformers for object detection. Li et al. proposed SegFormer \cite{xie2021segformer}, a simple, efficient, and robust semantic segmentation method.

Several studies have integrated Transformer-based methods into the field of crowd counting. TransCrowd \cite{liang2022transcrowd} is the earliest work that constructs the weakly supervised crowd counting from the perspective of sequence to counting. CCTrans \cite{tian2021cctrans} adopted a pyramid transformer as the backbone and designed a multi-scale regression head to alleviate scale variations. CTASNet \cite{chen2022counting} employed CNN to estimate low-density crowds, while leveraging Transformer for high-density crowds, and implemented an adaptive selection and fusion at the task level. Nevertheless, it failed to fully explore the deep intersections and fusion between these two architectures. Qian et al. \cite{qian2022segmentation} proposed a multi-scale Transformer and used a segmentation-based attention module to obtain fine-grained features. SAANet \cite{wei2021scene} designed a deformer backbone to extract the features, aggregates multi-level features by a deformable transformer encoder, and introduced a count query in a transformer decoder to re-calibrate the multi-level feature maps. Phan et al \cite{gao2020pccnet}. employed CNN and Transformer architectures as the foundational network structures to develop a crowd counting framework that integrates density estimation and object detection.


These Transformer-based methods employ self-attention mechanism to capture global dependencies. However, they often lack the ability to extract local detail information, which is crucial for accurately identifying targets within complex backgrounds. In contrast, our approach integrates global self-attention with CNN-based local attention, enabling the simultaneous modeling of both global and local dependencies. This dual-focus strategy effectively leverages global contextual information and local detail information, thereby significantly enhancing the ability to accurately discern targets in complex scenes.

\subsection{Deformable Mechanism}

Deformable convolution \cite{dai2017deformable,zhu2019deformable} is an effective mechanism that dynamically adjusts the shape of the convolutional kernel to better adapt to the features of target regions, effectively capturing non-uniform features of the target. Some researchers have incorporated the concept of deformable convolution into Transformer structures. Deformable DETR \cite{zhu2020deformable} is the first to introduce the deformable idea into the Transformer architecture. Li et al. \cite{xia2022vision} proposed a deformable attention mechanism to adaptively perceive spatial structures. Liu et al. \cite{wang2023internimage} draw inspiration from the foundational principles of Transformer, introducing DCNv3, which achieves a new performance record on the COCO dataset. Concurrently, designing novel deformable convolutions based on the characteristics of the task is also a focal point of research. Qi et al. \cite{qi2023dynamic} designed dynamic snake convolution based on the characteristics of tubular structures to enhance the perception of elongated structures. Zhang et al. \cite{zhong2022improved} analyzed the shape characteristics of human heads and proposed a Normed-Deformable Convolution. Zuo et al. \cite{zhang2023difference} produced a deformable attention module that combines a sparse spatial sampling strategy with long-range relationship modeling capability. Hence, given the significant scale variations and continuous changes characteristic in crowd counting tasks, we propose an input-dependent deformable convolution. By simultaneously learning convolution kernel offsets and weights based on input features, this approach enhances the perception of scale-varying targets.

\section{Methodology} \label{sec:method}
In this section, we first introduce the strong baseline and then introduce the proposed RCCFormer and the loss function.

\subsection{Strong Baseline}
\label{sec:strong baseline}
Most methods adopt the last layer of the backbone for subsequent feature extraction. Nevertheless, the features extracted by merely relying on the last layer are inadequate. Thus, the straightforward approach is to extract features from multiple stages of the backbone. Previous methods \cite{tian2021cctrans,liu2023multi} mainly employ concatenation or element-wise addition for fusion. As depicted in Fig. \ref{fig_1_2} (a), \textbf{Element-wise addition} implies aggregating by upsampling the deep features and adding pixel by pixel with the shallow features. While it can reduce complexity, it might fail to handle semantic discrepancies among features. \textbf{Concatenation} (see Fig. \ref{fig_1_2} (b)) indicates aggregating feature maps from different stages along the channel dimension. However, the simple concatenation method still cannot effectively fuse features selectively. To address the aforementioned issues, we propose a novel \textbf{Multi-level Feature Fusion Module} for fusing backbone features via the cross attention mechanism \cite{vaswani2017attention} (see Fig. \ref{fig_1_2} (c)). This operation assists the model in selectively and deliberately focusing on crucial information from different input features. Subsequently, we added it to the vanilla crowd counting network as a \textbf{Strong Baseline}. Then, we will introduce this module in detail.

\begin{figure}[!h]
	\centering
	\includegraphics[width=3.5in]{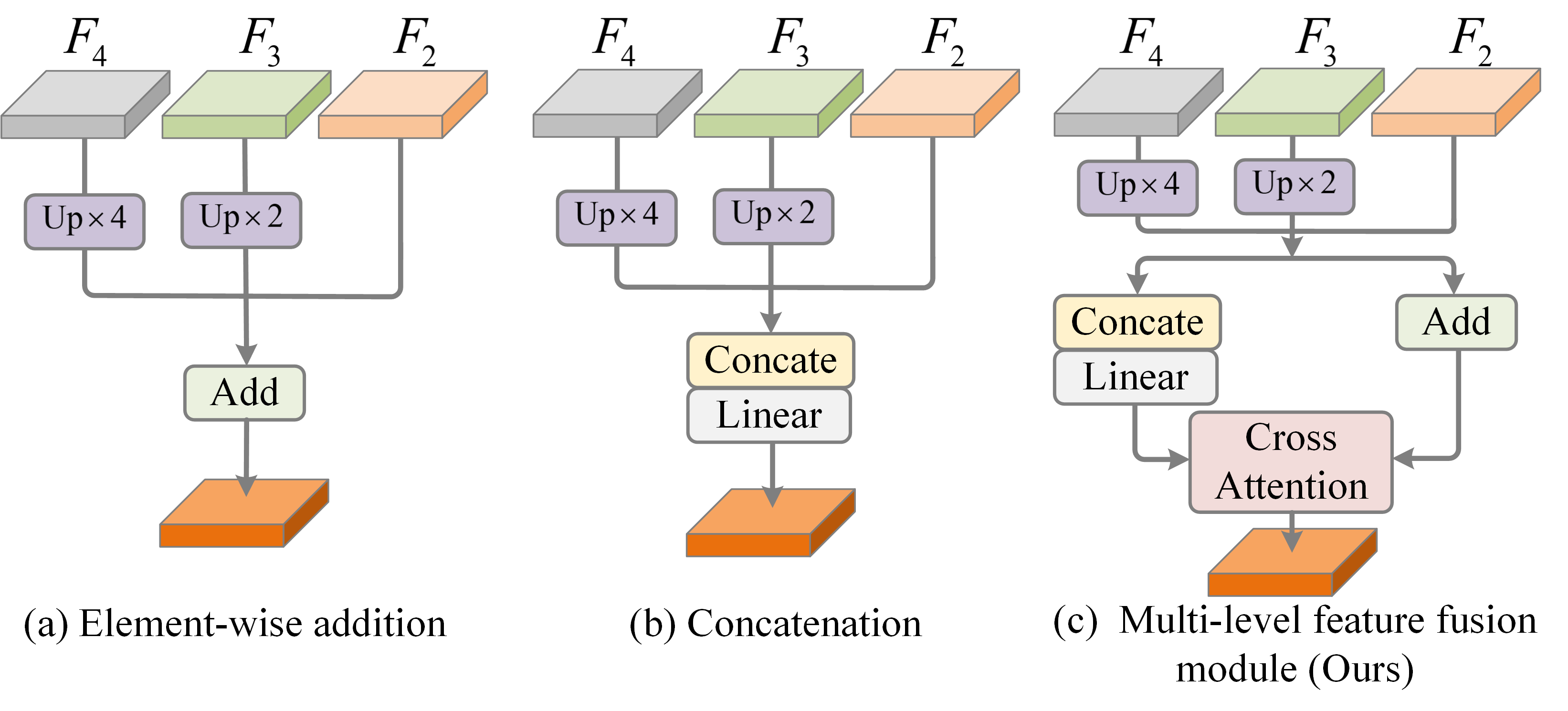}
	\caption{Different multi-level fusion methods. (a) element-wise addition; (b) concatenation; (c) our strong baseline. }
	\label{fig_1_2}
\end{figure}

\begin{table}[!h]
	\centering
	\caption{Comparison with the element-wise addition, concatenation and strong baseline on Part\_A and UCF\-QNRF datasets.}
	\renewcommand\arraystretch {1.2}
	\setlength\tabcolsep{1.8mm}%
	\label{table1}
	\begin{tabular}{lllllllll}
		\toprule[1pt]
		Method     & Part\_A &       &  & QNRF   &     \\ \cline{2-3} \cline{5-6}
		& MAE    & MSE    &  & MAE    & MSE  \\ \hline
		Element-wise addition       & 53.5  & 82.5  &  & 83.5   & 156.6 \\
		Concatenation     & 53.9   & 91.2  &  & 83.8   & 146.6 \\
		\rowcolor{gray!25} Strong baseline(Ours)   & \textbf{52.6}   & \textbf{82.2}   &  & \textbf{81.4}    & \textbf{146.2} \\
		\bottomrule[1pt]
	\end{tabular}

        \label{strong baseline}
\end{table}
To be specific, we first use concatenation and element-wise addition simultaneously for initial fusion and then introduce cross-attention \cite{vaswani2017attention} for fine fusion. As shown in Fig. \ref{fig_1_2} (c), for the output features $\left \{ F_2, F_3, F_4 \right \} $, the feature size is firstly unified to one-eighth of the input image resolution through the linear layer and upsampling layer, yielding features $\left \{ F_{2}^{'}, F_{3}^{'}, F_{4}^{'} \right \} $. In the concatenation branch, we employ concatenation and fuse different-level features using a linear layer to obtain $F_{c}$. In the addition branch, an element-wise addition is employed to obtain $F_{a}$. The detailed process is as follows:
\begin{equation}
    \begin{split}
        \label{deqn_ex1a}
        F_{c} &= \text{Linear}(\text{Concat}(F_{2}^{'}, F_{3}^{'}, F_{4}^{'})) \\
        F_{a} &= F_{2}^{'} + F_{3}^{'} + F_{4}^{'}
    \end{split}
\end{equation}
where $\text{Concat}$ is the concatenation and $\text{Linear}$ denotes the linear layer. Then, the concatenated feature $F_{c}$ is treated as query ($Q$), and the added feature $F_{a}$ is regarded as key ($K$) and value ($V$). We leverage a cross-attention to interact with them and obtain the final fusion result, denoted as $F_{f}$.
\begin{equation}
    \label{deqn_ex1_1a}
    F_{f} = \text{CrossAttn}(F_{c}, F_{a}, F_{a})
\end{equation}
where $\text{CrossAttn}(\cdot, \cdot,\cdot)$ is the cross-attention.

We validated the effectiveness of our method on two crowd counting datasets named Part\_A and UCF-QNRF. As shown in Table \ref{strong baseline}, our proposed strong baseline outperforms two traditional feature fusion methods, indicating that our approach extracts more robust features suitable for subsequent processing.

\begin{figure*}[!t]
	\centering
	\includegraphics[width=7.25in]{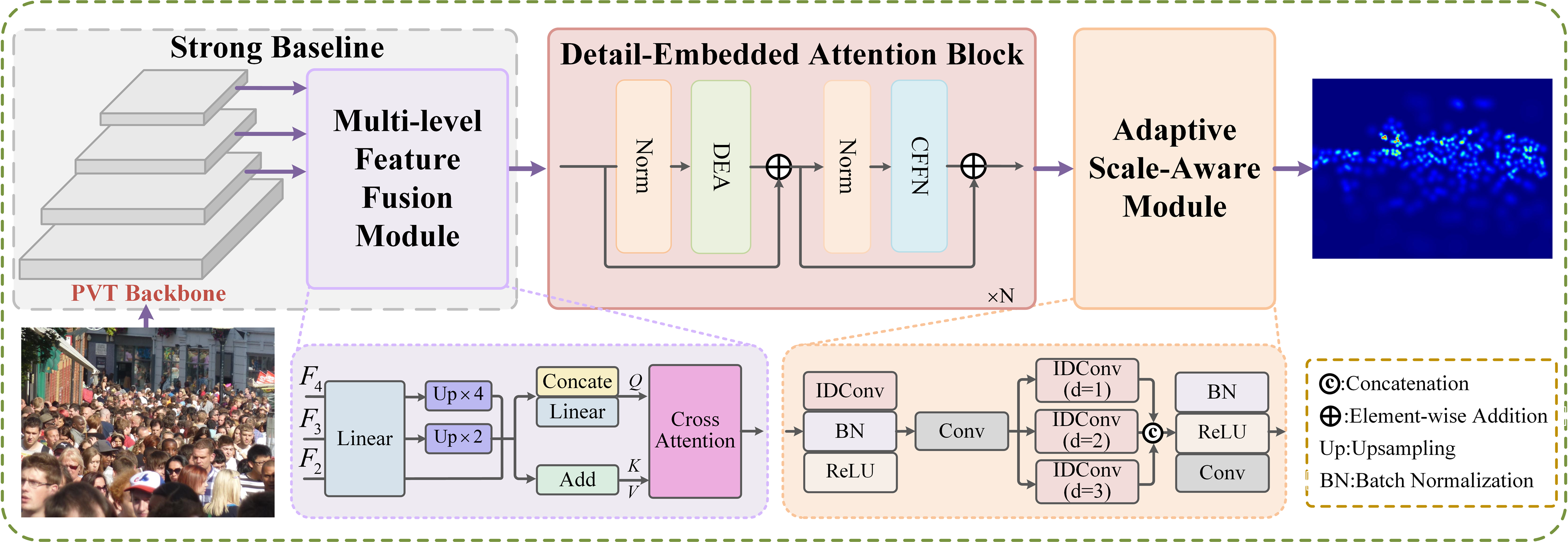}
	\caption{The overall framework of the proposed RCCFormer. It consists of three components: a strong baseline which contains pvt backbone and multi-level feature fusion module (MFFM), detail-embedded attention block (DEAB), and adaptive scale-aware module (ASAM). Among these, the MFFM integrates multi-level information; the DEAB extracts both global and local information for better suppression of background noise; the ASAM enhances the network's perceptual ability to large-scale variations. Some components are omitted for simplification. Please refer to the text for details.}
	\label{fig_2}
\end{figure*}
\subsection{RCCFormer}

Based on the Strong baseline, we further proposed RCCFormer. The overall framework of the proposed architecture is shown in Fig.  \ref{fig_2}. It consists of four main components: 1) Backbone network, where the Transformer network PVT-v2 is used to extract multi-level features $\left \{ F_2,F_3,F_4 \right \} $; 2) \textbf{Multi-level Feature Fusion Module} (MFFM), which utilizes concatenation, element-wise addition, and cross-attention to achieve better multi-level information fusion; 3) \textbf{Detail Embedded Attention Block} (DEAB), which fully exploits global and local contextual information to amplify foreground targets and suppress background noise; 4) \textbf{Adaptive Scale-Aware Module} (ASAM), where IDConvs with diverse dilation rates are employed to enhance the adaptive perception capability for large-scale objects. The details of DEAB and ASAM are described below.

\subsubsection{Detail-Embedded Attention Block}
When confronted with complex background challenges in crowded scenes, although Transformers can leverage self-attention mechanism to model global pixel-wise correlations, they lack positional awareness and fail to effectively attend to local details which are also important for understanding complex scenes. Therefore, we propose Detail-Embedded Attention Block (DEAB), an effective module for modeling global-local context information. Initially, through query ($Q$) and key ($K$) interactions, we obtain dependencies between global pixels and use it as global attention to perform matrix multiplication with value ($V$) to capture context information. Meanwhile, inspired by \cite{lau2024large}, we design local attention to extract local fine-grained representations. Specifically, a convolutional module with the local receptive field is employed to extract local dependencies, and its output is used as attention, which then performs element-wise multiplication with the $V$ to generate the final output. In this way, the dynamic property can be obtained which is similar to those of the self-attention mechanism, effectively bridging the inherent semantic gap between CNNs and Transformers. Finally, the learnable manner is adopted to fuse the global and local context information selectively.

The overall structure of DEA is shown in Fig. \ref{fig_3}. It can be seen that this module has two branches: the local attention branch and the global self-attention branch. For the global self-attention branch, given the input $X\in \mathbb{R}^{H\times W\times C}$, we reshape it into tokens $X_{t}\in \mathbb{R}^{HW\times C}$ and then project it into $Q\in \mathbb{R}^{HW\times C}$, $K\in \mathbb{R}^{HW\times C}$ and $V\in \mathbb{R}^{HW\times C}$:
\begin{equation}
	\begin{split}		
		\label{deqn_ex3a}
		Q=X_{t}W^Q,K=X_{t}W^K,V=X_{t}W^V
	\end{split}
\end{equation}
where $W^K,W^V$ and $W^Q\in \mathbb{R}^{C\times C}$ are learnable projection matrixes. Subsequently, we split $Q$, $K$, and $V$ into $N$ heads along the channel dimension: $Q = [Q_{1}, ... , Q_{N} ], K = [K_{1}, ..., K_{N}]$, and $V = [V_{1}, ..., V_{N}]$, respectively. The dimension of each head is $d_{h}=\frac{C}{N} $. Thus, for $head_{j}$, the self-attention is :
\begin{equation}
    \label{deqn_ex4a}
    Attn_{ga_j} = \text{Softmax} \left( \frac{Q_{j} K_{j}^T}{\sqrt{d_h}} \right) \otimes V_{j}
\end{equation}

\begin{figure}[!t]
	\centering
	\includegraphics[width=3.in]{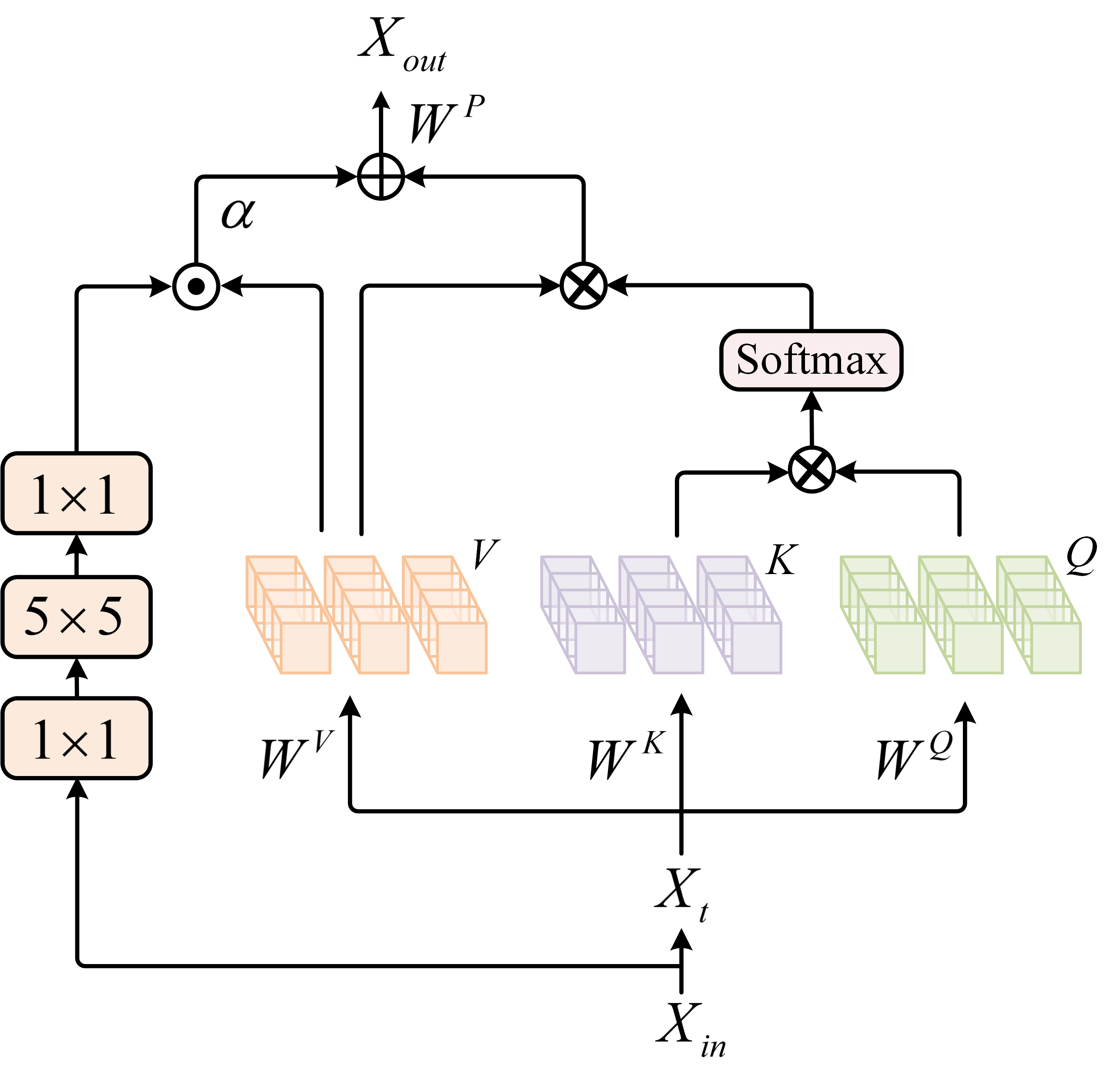}
	\caption{Illustration of the proposed Detail-Embedded Attention (DEA).}
	\label{fig_3}
\end{figure}

For the local attention branch, given the same input $X\in \mathbb{R}^{H\times W\times C}$, we first feed it into $1\times 1$ convolutional layer to projection. To better model the local detail information, we use one $5\times 5$ depth-wise convolution to leverage local detail information and employ a $1\times 1$ convolution to aggregate information across channels. Finally, to align with the multi-head features obtained from the global branch, we reshape the output features to $X\in \mathbb{R}^{N\times (H\times W)\times \frac{C}{N}}$. Similarly, for $head_{j}$, the local-attention is:

\begin{equation}
    \label{deqn_ex5a}
    Attn_{la_j} = \text{Conv}_{1 \times 1} \left( \text{Conv}_{5 \times 5} \left( \text{Conv}_{1 \times 1} \right) \right) \odot V_{j}
\end{equation}

Finally, to enable the fusion of features in a learnable manner, we introduce a learnable parameter acting on local attention. So the output of the overall attention is:
\begin{equation}
	\begin{split}		
		\label{deqn_ex6a}
		Attn_j = Attn_{ga_j} + \alpha Attn_{la_j}
	\end{split}
\end{equation}
where $\alpha$ is a learnable parameter which is initialized to 0.6.

Then, the detail-embedded attention is:
\begin{equation}
    \label{deqn_ex7a}
    DEA = \text{Concat}(Attn_{1}, \ldots, Attn_{N}) W^P
\end{equation}
where $W^P$ is also a learnable projection matrix.

After that, we use DEA as the core component and combine it with LayerNorm and convolutional feed-forward neural network (CFFN)\cite{wang2022pvt} which is used to aggregate features to build a DEAB which is shown in the Fig. \ref{fig_2}. This arrangement can be expressed as: 

\begin{equation}
    \begin{aligned}
        \hat{X} &= \text{DEA}(\text{LN}(X_{in})) + X_{in} \\
        X_{out} &= \text{CFFN}(\text{LN}(\hat{X})) + \hat{X}
    \end{aligned}
    \label{deqn_ex8a}
\end{equation}
where $X_{in}$ is input feature, $\text{LN}$ is layer normalization, and $\text{CFFN}$ denotes the convolutional feed-forward neural network.

\subsubsection{Adaptive Scale-Aware Module}
In the previous discussion, the joint modeling of global and local dependencies has demonstrated its effectiveness in tackling complex background issues. However, the challenge of scale variation in crowd counting remains largely unresolved. Traditional methods frequently utilize convolutions with different kernel sizes, e.g., $3\times 3$, $5\times 5$, and $7\times 7$, or vary dilation rates to adapt to scale changes. Nonetheless, these methods are constrained by the fixed weights and sizes of convolutional kernels, limiting their adaptability to scale variations. Some others \cite{bai2020adaptive,yan2021crowd} introduce deformable convolution and design adaptive dilation convolution to alleviate the shortcomings of vanilla convolution for crowd counting tasks, but only consider the geometric shape of the head and do not fully capture the head instance. 
\begin{figure}[!t]
	\centering
	\includegraphics[width=3.5in]{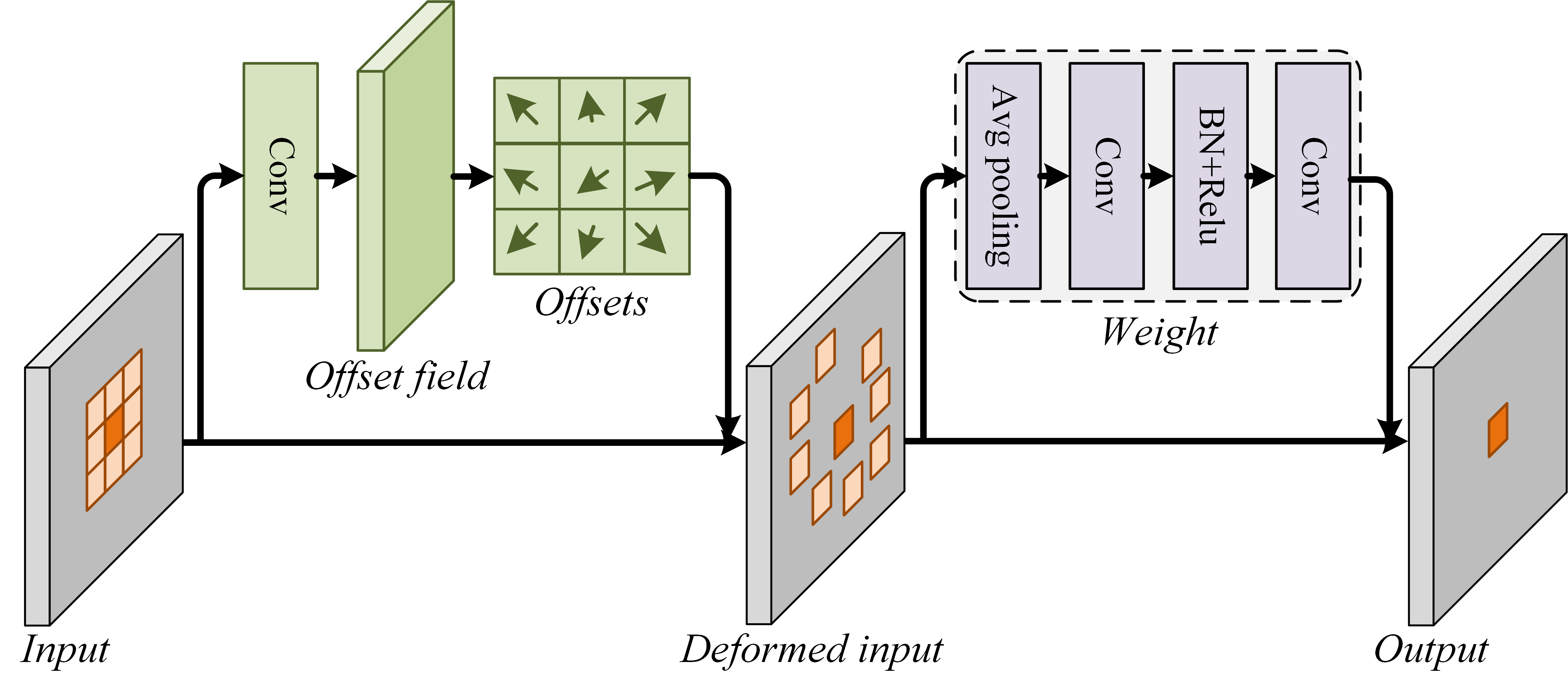}
	\caption{Illustration of the proposed Input-dependent Deformable Convolution.}
	\label{fig_4}
\end{figure}
To this end, we propose an input-dependent deformable convolution and build an adaptive scale-aware module, the details are as follows.

{\bf{Input-dependent Deformable Convolution:}} inspired by the dynamic weight adjustment mechanism in self-attention, we propose an Input-Dependent Deformable Convolution named IDConv. This convolution adaptively learns the deformation parameters and weight coefficients of convolutional kernels, enabling these kernels to dynamically adjust their shapes and values in response to input data. This capability can provide strong robustness in crowd counting tasks, where the target scale and shape can vary significantly. Specifically, as illustrated in Fig. \ref{fig_4}, our IDConv adaptively adjusts the offset of convolution sampling points and dynamically generates input-aware weights during the feature extraction process. Formally, the IDConv can be represented as
\begin{equation}
	\begin{split}
		\label{deqn_ex9a}
		y(p_0)=\sum_{p_n\in R}^{} w(p_n)\times x(p_0+p_n+\Delta p_n)
	\end{split}
\end{equation}
where $R$ represents the local receptive field, e.g., $R={(-1,-1),(-1,0),...,(0,1),(1,1)}$ when convolution kernel is $3\times 3$, $p_0$ denotes the central position of  $R$, $p_n$ represents the relative position of each value from
$R$ to $p_0$, and $w(p_n)$ is a learnable weight, that is expressed as (\ref{deqn_ex10a}). $\bigtriangleup p_n$ is a learnable offset which is
generated by the $3\times 3$ convolution, and define $\bigtriangleup p_n$ as follows: 
\begin{equation}
    \begin{aligned}
        \label{deqn_ex10a}
        \Delta p_n &= \text{Conv}_{3 \times 3}(x) \\
        w_{gap} &= \text{GAP}(x(p_0 + p_n + \Delta p_n)) \\
        w(p_n) &= \text{Conv}_{1 \times 1} \left( \text{ReLU} \left( \text{BN} \left( \text{Conv}_{1 \times 1}(w_{gap}) \right) \right) \right)
    \end{aligned}
\end{equation}
where $\text{Conv}_{3\times 3}$ represents $3\times 3$ convolution, $\text{Conv}_{1\times 1}$ represents $1\times 1$ convolution, $\text{BN}$ is batch normalization, and $\text{ReLU}$ is relu activation function. The $\text{GAP}$ is a global average pooling operator.

{\bf{Adaptive Scale-Aware Module:}} the IDConv is used as the basic component to build an adaptive scale-aware module as shown in Fig. \ref{fig_2}. For the input feature $F_{in}\in \mathbb{R} ^{H\times W\times C} $, we initially use a $3\times 3$ IDConv with batch normalization and ReLU activation function to capture the local spatial information, and though $1\times 1$ convolution to reduce channels to the half of the origin image, resulting in $F_{m}\in \mathbb{R} ^{H\times W\times C} $. Then, designing three parallel dilated IDConv operations, which dilations = $\left \{1,2,3  \right \} $, to capture a large-scale range of the object. After that, the batch normalization and ReLU activation function are also used to provide normalization and nonlinearity, followed by a $1\times 1$ convolution to obtain the output result $F_{out}\in \mathbb{R} ^{H\times W\times C}$:
\begin{equation}
    \begin{aligned}
        F_{m} &= \text{Conv}_{1 \times 1} \left( \text{ReLU} \left( \text{BN} \left( \text{IDConv}_{3 \times 3} (F_{in}) \right) \right) \right) \\
        F_{c} &= \text{Concat} \left( \text{IDConv}_{3 \times 3, d} (F_{m}) \right)_{d \in \{1, 2, 3\}} \\
        F_{out} &= \text{Conv}_{1 \times 1} \left( \text{ReLU} \left( \text{BN} (F_{c}) \right) \right)
    \end{aligned}
\end{equation}

\subsection{Loss Function}
In our method, the loss function is based on DM-count \cite{wang2020distribution} which considers crowd counting as a distribution matching problem and measures the similarity of the predicted density map and ground truth. Compared to traditional Gaussian density maps, this loss accurately captures real-world crowd distribution, particularly in dense regions. It consists of three components: the counting loss $L_C$ , the optimal transport loss $L_{OT}$ and the total variation loss $L_{TV}$, which are formulated as follows:
\begin{equation}
	\label{deqn_ex14a}
	L_{count} = L_C(C^\prime ,C) + \lambda _1L_{OT}(D^\prime ,D) + \lambda _2L_{TV}(D^\prime ,D)
\end{equation}
where $C^\prime$ is the predicted count and $C$ is ground truth count; $D^\prime $ and $D$ represents the predicted density map and the ground truth. $\lambda _1$ and $\lambda _2$ are the weights of losses to realize a better mix of these losses which are set to 0.1 and 0.01, respectively.

\section{Experiments} \label{sec:experiments}
In this section, we first introduce the crowd counting dataset and experimental setting. Then, comparison experiments and ablation studies are carried out to verify the performance of our method.
\subsection{Datasets}
We conduct extensive experiments on four most commonly used crowd counting datasets: ShanghaiTech Part\_A and Part\_B \cite{zhang2016single}, UCF-QNRF \cite{idrees2018composition} and NWPU-Crowd \cite{wang2020nwpu}, which are described as follows:

{\bf{ShanghaiTech dataset}} consists of Part\_A and Part\_B, comprising a total of 1198 crowd images annotated by 330,165 people. Part\_A includes 482 crowded internet images, with 300 images allocated for training and 182 for testing. Part\_B consists of 716 sparse images captured at a fixed size of 768×1024, with 400 images designated for training and 316 for testing.

{\bf{UCF-QNRF dataset}} contains 1535 high-resolution images, with the number of people per image ranging from 49 to 12865, totaling 1,252,542 people. It owns the characteristics of diverse density, complex background, and distorted perspectives, which is extremely challenging.

{\bf{NWPU-Crowd dataset}} is a large-scale dataset that consists of 5109 images, 3109 for training, 500 for validating, and 1500 for testing. The number of people in a single image ranges from 0 to 20033 in various diversity of scenes with density variation and scale variation. In addition, 351 negative images are introduced to this dataset and it is more challenging for counting. According to density level, the testing set is divided into five fine-grained subsets, (1) S0: images of negative samples, (2) S1: images containing $1\sim100$ people, (3) S2: images with $101\sim500$ people, (4) S3: images with $501\sim5000$ people and (5) S4: images contain more than 5000 people.

\subsection{Experiment Settings}
{\bf{Implement details:}}
Our approach adopts the backbone of PVTv2-B3 which is pre-trained on ImageNet-1K. For the data augmentation, we use random cropping and random horizontal flipping during the training stage. The image size is set to $512\times 512$ for ShanghaiTech Part\_B and QNRF dataset, and $384\times 384$ for the NWPU-Crowd. As some images in ShanghaiTech Part\_A are in low resolution, the size changes to $256\times 256$. Additionally, the batch size is 6 for all datasets, except for ShanghaiTech Part\_A, which is set to 16. We adopt AdamW as the optimizer with the learning rate set to $1e^{-5}$ and weight decay set to $1e^{-4}$. All experiments are conducted on a single NVIDIA RTX4090 GPU using the PyTorch framework.

{\bf{Evaluation Metrics:}} For the ShanghaiTech and QNRF datasets, we use Mean Absolute Error (MAE) and Mean Squared Error (MSE) as metrics to evaluate the performance of different methods. However, for the NWPU dataset, following \cite{wang2020nwpu}, we have introduced the additional evaluation metric NAE (mean Normalized Absolute Error). These metrics are defined as follows:
\begin{equation}
	\label{deqn_ex15a}
	MAE = \frac{1}{N}\sum_{i=1}^{N}\left | z_i-z_i^{gt} \right |
\end{equation}
\begin{equation}
	\label{deqn_ex16a}
	MSE = \sqrt{\frac{1}{N} \sum_{i=1}^{N} (z_i - z_i^{gt})^2}    
\end{equation}
\begin{equation}
	\label{deqn_ex17a}
	NAE = \frac{1}{N}\sum_{i=1}^{N}\frac{\left | z_i - z_i^{gt} \right |}{z_i^{gt}} 
\end{equation}
where $N$ is the number of test images, $z_i$ and $z_i^{gt}$ are the estimated number of people and the ground truth in the $i$-th images. It is noteworthy that the NWPU-Crowd has a number of negative samples with zero annotation, they are excluded during the calculation of $NAE$ to avoid zero denominators. 

\begin{table*}
	\centering     
	\caption{Comparison with the state-of-the-art methods on Part\_A, Part\_B and UCF-QNRF datasets. The best results are highlighted in red, and the second-best results are highlighted in blue.}
	\renewcommand\arraystretch {1.2}
	\definecolor{MyRed}{RGB}{255,0,0}
	\definecolor{MyBlue}{RGB}{0,0,255}
	\setlength\tabcolsep{3mm}%
	\label{table1-1}
	\begin{tabular}{llllllllllll}
		\toprule[1pt]
		Method         & Part\_A &       &  & Part\_B &      &  & UCF-QNRF &     \\ \cline{2-3} \cline{5-6} \cline{8-9}
	           & MAE $\downarrow$      & MSE $\downarrow$    &          & MAE $\downarrow$    & MSE $\downarrow$  &  & MAE $\downarrow$      & MSE $\downarrow$   \\ \hline
		MCNN \cite{zhang2016single}          & 110.2  & 173.2 &  & 26.4   & 41.3 &  & 277.0    & 426.0 \\
		DM-Count \cite{wang2020distribution}    & 59.7  & 95.7  &  & 7.4    & 11.8 &  & 85.6     & 148.3 \\
		P2PNet \cite{song2021rethinking}        & 52.8   & 85.1  &  & 6.3  & 9.9  &  & 85.3     & 154.5 \\
		ChfL \cite{shu2022crowd}		   & 57.5   & 94.3  &  & 6.9    & 11.0 &  & 80.3     & \color{MyBlue}{137.6} \\
		STNet \cite{wang2022stnet}          & 52.8   & 83.6  &  & \color{MyBlue}{6.2}    & 10.3 &  & 87.9     & 166.4 \\
		GGANet \cite{guo2023object}     & 62.0   & 110.7  &  & 7.4    & 13.1  &  & 91.9     & 156.8 \\ 
        CAAPN  \cite{liu2024CAAPN}          & 54.6   & 100.5  &  & \color{MyRed}{5.9}    & 10.6  &  & 87.5     & 138.5 \\ \hline
		TransCrowd \cite{liang2022transcrowd}   & 66.1   & 105.1 &  & 9.3    & 16.1 &  & 97.2     & 168.5 \\
		CCTrans \cite{tian2021cctrans}      & 52.3   & 84.9  &  & \color{MyBlue}{6.2}    &9.9  &  & 82.8     & 142.3 \\
		CLTR \cite{liang2022end}         & 56.9   & 95.2  &  & 6.5    & 10.6 &  & 87.3     & 142.4 \\
        CTASNet \cite{chen2022counting}      & 54.3   & 87.8  &  & 6.5    & 10.7 &  & 80.9     & 139.2 \\
        PET \cite{liu2023point}          & \color{MyBlue}{49.3}   & 78.8  &  & \color{MyBlue}{6.2}    & \color{MyBlue}{9.7}  &  & \color{MyBlue}{79.5}     & 144.3 \\
        CF-former \cite{jiang2024CF-former}          & 50.8   & \color{MyBlue}{77.5}  &  & 6.6    & 10.3  &  & 82.4     & 140.5 \\
        DEO-Net \cite{gao2020pccnet}           & 54.2   & 85.3 &  & \color{MyBlue}{6.2}   & \color{MyRed}{9.6} &  & 83.1  & 141.5 \\
		\rowcolor{gray!25} RCCFormer (Ours)                & \color{MyRed}{48.3}   & \color{MyRed}{72.1}  &  & 6.6    & 10.4 &   & \color{MyRed}{77.6}     & \color{MyRed}{133.9} \\ 
		\bottomrule[1pt]
	\end{tabular}
\end{table*}

\begin{table*}
	\centering
	\caption{Comparison with the state-of-the-art methods on NWPU-CROWD dataset. The best results are highlighted in red, and the second-best results are highlighted in blue. }
	\renewcommand\arraystretch {1.2}
	\definecolor{MyRed}{RGB}{255,0,0}
	\definecolor{MyBlue}{RGB}{0,0,255}
	\setlength\tabcolsep{3mm}%
	\label{table2}
	\begin{tabular}{llll|ll}
		\toprule[1pt]
		\multirow{2}{*}{Method}  & \multicolumn{3}{c|}{Overall} & \multicolumn{2}{c}{Scene Level(MAE)}                                \\ \cline{2-4} \cline{5-6} 
		& MAE $\downarrow$      & MSE $\downarrow$     & NAE $\downarrow$     & \multicolumn{1}{l|}{Avg. $\downarrow$}   & \multicolumn{1}{c}{S0--S4 $\downarrow$}            \\ \hline
		MCNN \cite{zhang2016single}                                  & 232.5    & 714.6   & 1.063   & \multicolumn{1}{l|}{1171.9} & 356.0 / 72.1 / 103.5 / 509.5 / 4818.5 \\
		CSRNet \cite{li2018dilated}                                  & 121.3    & 387.8   & 0.604   & \multicolumn{1}{l|}{522.7}  & 176.0 / 35.8 / 59.8 / 285.8 / 2055.8  \\
		CAN \cite{liu2019context}                                     & 106.3    & 386.5   & 0.295   & \multicolumn{1}{l|}{612.2}  & 82.6 / 14.7 / 46.6 / 269.7 / 2647.0   \\
		DM-Count \cite{wang2020distribution}                           & 88.4     & 388.6   & 0.169   & \multicolumn{1}{l|}{498.0}  & 146.7 / 7.6 / 31.2 / 228.7 / 2075.8   \\
		UOT \cite{ma2021learning}                                  & 87.8     & 387.5   & 0.185   & \multicolumn{1}{l|}{566.5}  & 80.7 / 7.9 / 36.3 / 212.0 / 2495.4   \\
        AutoScale \cite{xu2022autoscale}     & 94.2     & 388.2   & 0.226   &\multicolumn{1}{l|}{608.2}  & 81.4 / 11.4 / 38.2 / 226.1 / 2683.7   \\
		ChfL \cite{shu2022crowd}                                   & 76.8     & 343.0   & 0.171   & \multicolumn{1}{l|}{470.0}  & 56.7 / 8.3 / 32.1 / 195.1 / 2058.0    \\
        CAAL \cite{meng2023caal}                                    & 76.4     & \color{MyBlue}{327.4}   & 0.182   & \multicolumn{1}{l|}{470.0}  & 27.9 / 8.2 / 37.3 / 189.7 / 2075.3    \\ \hline
		TransCrowd \cite{liang2022transcrowd}                             & 117.7    & 451.0   & 0.244   & \multicolumn{1}{l|}{737.8}  & 69.3 / 12.8 / 45.9 / 308.8 / 3252.2   \\
        MAN \cite{lin2022boosting}                                    & 76.5     & \color{MyRed}{323.0}   & 0.170   & \multicolumn{1}{l|}{464.6}  & 43.3 / 8.5 / 35.3 / 190.9 / {\color{MyBlue}{2044.9}}    \\
        CLTR \cite{liang2022end}                           & \color{MyBlue}{74.4}     & 333.8   & \color{MyBlue}{0.165}   & \multicolumn{1}{l|}{\color{MyBlue}{423.3}}  & {\color{MyRed}{4.2}} / {\color{MyBlue}{7.3}} / {\color{MyBlue}{30.3}} / {\color{MyBlue}{185.5}} / 2434.8    \\
         CTASNet \cite{chen2022counting}                    & 94.4     & 357.6   & 0.213      & \multicolumn{1}{l|}{497.3}    & 52.1 / 10.4 / 42.5 / 264.1 / 2117.3 \\
        PET \cite{liu2023point}                    & \color{MyBlue}{74.4}     & 328.4   & 0.193      & \multicolumn{1}{l|}{504.4}    & 41.4 / 10.7 / 32.2 / {\color{MyBlue}{170.1}} / 2267.9 \\
		\rowcolor{gray!25} RCCFormer (Ours)                                            & \color{MyRed}{74.3}     & 332.7  & \color{MyRed}{0.160}   & \multicolumn{1}{l|}{\color{MyRed}{401.0}}  & 82.2 / {\color{MyRed}{6.9}} / {\color{MyRed}{29.9}} / 201.8 / \color{MyRed}{1684.0}      \\
		\bottomrule[1pt]
	\end{tabular}
\end{table*}

\subsection{Comparison with State-of-the-art Methods}
We here conducted a series of qualitative experiments on these four datasets to verify the effectiveness of the proposed method. Notably, the majority of existing methods rely on CNN and Transformer architectures. 

From Table \ref{table1-1}, our model outperforms existing methods on the ShanghaiTech Part\_A dataset, achieving the best results with an MAE of 48.3 and MSE of 72.1. Compared to CNN-based methods, our model reduces MAE by 4.5 and MSE by 11.5 from the second-best result. Particularly, compared to existing Transformer-based approaches, our model reduces MAE by 1.0 and MSE by 5.4. For the ShanghaiTech Part\_B dataset, our method shows a slight gap compared to the best result. The reason is the sparse distribution in this dataset, which results in heads occupying a relatively larger proportion of the images and containing significant detail information. This necessitates a heightened focus on local features for accurate counting. However, our model is based on Transformer, which primarily focuses on global contextual information and only partially attends to local details, resulting in insufficient capture of the local features. On the UCF-QNRF dataset, our method also achieves the best result among the Transformer-based and CNN-based methods. Compared to the second-best result, our model shows a decrease of 1.9 in MAE and 3.7 in MSE, which demonstrates the enhanced performance of our model under large-scale variations and complex scenes.

To further investigate the performance of our RCCFormer on large-scale datasets, we conducted experiments on the NWPU-Crowd dataset. Compared to the previous three datasets, this dataset exhibits higher complexity in scene settings. The results in Table \ref{table2} demonstrate the effectiveness of our method. Our proposed RCCFormer achieves the best performance in MAE and NAE, with values of 74.3 and 0.160, respectively. However, because of the diversity in scene distribution within the NWPU dataset, the model cannot adapt well to all scenes, resulting in relatively high MSE. Notably, in the results for different density levels, it is observed that our model significantly outperforms previous methods on S4, which contains more than 5000 heads per image, indicating the superior performance of our model in high-density scenes.

The experimental analysis validates that our model excels in counting not only on sparse and simple images but also exhibits strong robustness in large-scale, complex crowd scenes. It effectively addresses the challenges posed by scale variations and complex background interference.

\begin{figure}[!t]
	\centering
	\includegraphics[width=3.5in]{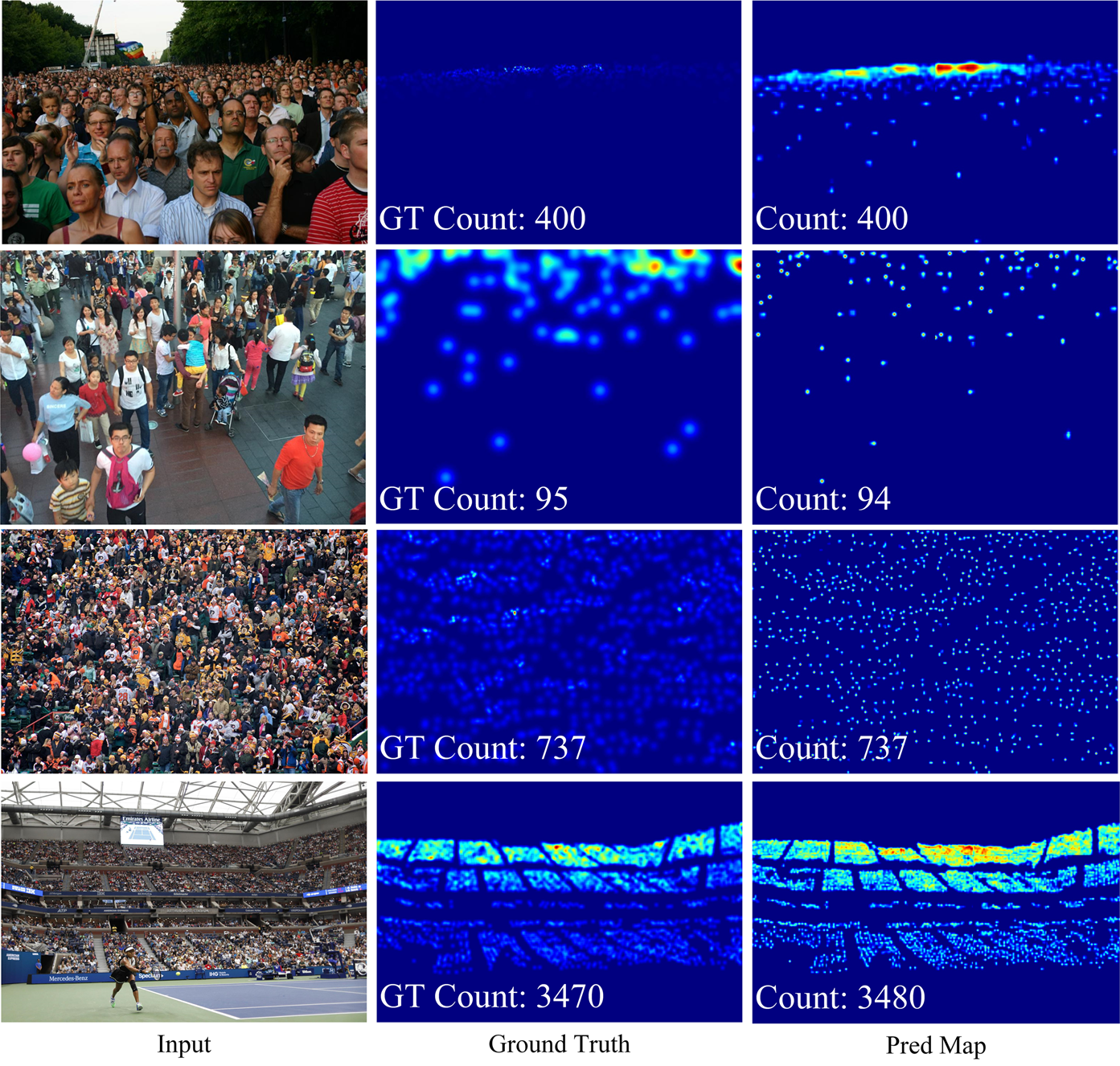}
	\caption{Visualization results on the four crowd-counting test datasets. From top to bottom, the images are from SHA, SHB, UCF-QNRF and NWPU-Crowd datasets, respectively.}
	\label{fig_5}
\end{figure}

\subsection{Visualization}
We conduct visual analysis on four datasets: ShanghaiTech Part\_A and Part\_B, QNRF, and NWPU. Some results are shown in Fig. \ref{fig_5}. The first row displays outcomes on ShanghaiTech Part\_A images, showcasing the model's roubustness in achieving precise counting even in large-scale and densely populated scenes. The second row illustrates outcomes on ShanghaiTech Part\_B images, verifying the model's robust performance in sparse scenes. Results on QNRF, depicted in the third row, underscore the model's effectiveness in addressing the challenges presented by complex and varied backgrounds. Finally, the last row portrays outcomes on NWPU, demonstrating the model's capability in both highly dense and complex scenes.

To further analyze the results with different methods, Fig. \ref{fig_7} presents visual comparisons of our RCCFormer with DM-Count \cite{wang2020distribution} and CCTrans \cite{tian2021cctrans}. The examples include images with dense crowds and simple backgrounds (first row), highly crowded images (second row), images with large-scale variations (third row), and images with dense crowds and complex backgrounds (fourth row). It can be seen that the proposed RCCFormer consistently performs well in all these scenarios, producing density maps that are closest to the ground truth compared to the other two methods. Notably, when faced with significant background interference, as shown in the fourth row, the compared methods either overestimate or underestimate the crowd count, whereas our model's predictions are close to the actual count. This demonstrates the robustness and effectiveness of our method.

\begin{figure*}[!t]
	\centering
	\includegraphics[width=7in]{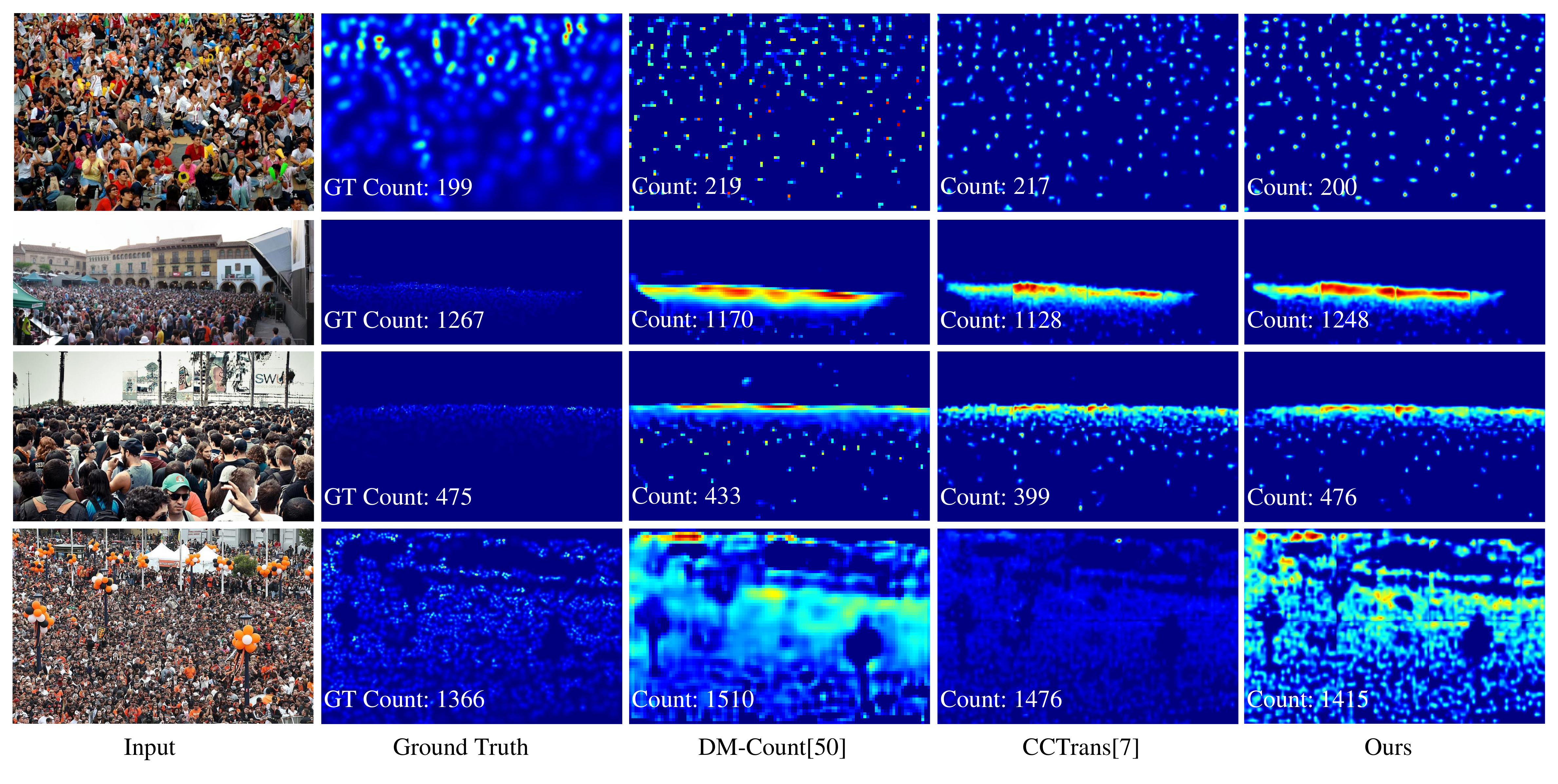}
	\caption{Visual comparison of predicted density maps generated by our method and DM-Count \cite{wang2020distribution}, CCTrans \cite{tian2021cctrans} under different scenes. It can be seen that our model consistently produces density maps closest to the ground truth.}
	\label{fig_7}
\end{figure*}

\subsection{Ablation Study}

\begin{table}
	\centering
	\caption{Ablation studies of each component in our method based on the part\_A dataset.}
	\renewcommand\arraystretch {1.2}
	\setlength\tabcolsep{3mm}%
	\label{table3}
	\begin{tabular}{llllll}
		\toprule[1pt]
		MFFM & DEAB & ASAM & MAE $\downarrow$   & MSE $\downarrow$   \\ \hline
		&        &     & 53.6 & 89.2 \\
		\Checkmark&        &     & 52.6 & 82.2 \\
		\Checkmark&        \Checkmark&     & 50.0 & 79.9 \\
		\Checkmark&        \Checkmark&     \Checkmark& \textbf{48.3} & \textbf{72.1} \\ \bottomrule[1pt]
	\end{tabular}
\end{table}

{\bf{Effectiveness of Various Components.}} To validate the effectiveness of the designed modules including MFFM, DEAB, and ASAM, we conduct ablation experiments on the ShanghaiTech Part\_A dataset, which the experimental results are shown in Table \ref{table3}.

We upsample the final layer of the backbone to one-eighth of the image resolution and use a $1\times 1$ convolution to output the density map as the baseline. From Table \ref{table3}, it can be observed that, compared to the baseline, the introduction of the MFFM results in a decrease of 1.0 in MAE and 7.0 in MSE. This module can effectively integrate low-level fine-grained information and high-level semantic information, consequently providing more accurate positional cues for head instances. With the addition of the  DEAB, the error further decreased, with MAE dropping to 50.0 and MSE decreasing to 79.9. It indicates that this block effectively achieves the separation of foreground and background, enhancing the extraction of foreground targets. Finally, with the implementation of the ASAM based on input-dependent deformable convolutions, MAE decreases by 1.7 and MSE decreases by 7.8, confirming the module's ability to perceive a large-scale range and to effectively extract features from targets of various scales.

Furthermore, we provide the illustration of density maps to visually analyze the impact of each module. Some visualization results are shown in Fig. \ref{fig_6}, where Fig. \ref{fig_6} (a) depicts the input image with complex background and large-scale range, and Fig. \ref{fig_6} (b) represents the ground truth. Fig. \ref{fig_6} (c)-(f) respectively show the output density maps after the addition of the baseline, MFFM, DEAB, and ASAM. As depicted in Fig. \ref{fig_6} (d), compared to the baseline in Fig. \ref{fig_6} (c), the introduction of the MFFM enables more accurate localization of head targets. Additionally, as shown in Fig. \ref{fig_6} (a), the red box indicates background interference such as tree branches, and the density map in Fig. \ref{fig_6} (d) is affected by the tree branches. In contrast, Fig. \ref{fig_6} (e) and (f) is hardly influenced by the complex background, indicating that the DEAB effectively filters out background noise. Finally, within the green box in Fig. \ref{fig_6} (a) containing two human head targets of different scales, Fig. \ref{fig_6} (d) and (e) miss the smaller-scale target and the density value of another target is relatively low. However, Fig. \ref{fig_6} (f) successfully detects both targets, demonstrating that the addition of the ASAM enhances the network's perception of target scales.
\begin{table}
	\centering
	\caption{Ablation studies of different fusion methods.}
	\renewcommand\arraystretch {1.2}
	\setlength\tabcolsep{5mm}%
	\label{table4}
	\begin{tabular}{lll}
		\toprule[1pt]
		Methods     & MAE $\downarrow$  & MSE $\downarrow$   \\ \hline
		Concate     & 53.9 & 91.2 \\
		Add         & 53.5 & 82.5 \\
		Concate+Add+Add & 53.8 & 86.0 \\
		Concate+Add+Concate & 52.8 & 83.9 \\
		\rowcolor{gray!25} MFFM (ours) & \textbf{52.6} & \textbf{82.2} \\ 
		\bottomrule[1pt]
	\end{tabular}
\end{table}

{\bf{Effect of Multi-level Feature Fusion Module.}} To validate the effectiveness of our proposed MFFM, we compare five different fusion methods. Among them, Concate+Add+Concate initially merges multi-level information separately through concatenation and element-wise addition and then utilizes concatenation to fuse the information between them. Similarly, Concate+Add+Add does the same but utilizes element-wise addition to finally fuse the information between them. Compared to using only Concate and Add, our MFFM method reduces MAE by 1.3 and 0.9, and MSE by 9 and 0.3, respectively. This indicates that further fusing the information between concatenation and element-wise addition using cross-attention can yield better multi-level fusion information. In comparison to Concate+Add+Concate and Concate+Add+Add, the experimental results of our MFFM method surpass both, demonstrating that cross-attention is a superior fusion method compared to Concate and Add.
\begin{table}
	\centering
	\caption{Ablation studies of DEA.}
	\renewcommand\arraystretch {1.2}
	 \setlength\tabcolsep{4mm}%
	\label{table6}
	\begin{tabular}{lll}
		\toprule[1pt]
		Methods     & MAE $\downarrow$   & MSE $\downarrow$   \\ \hline
		GSA     & 51.3 & 81.7 \\
		GSA + Local Convolution & 50.8 & 86.3 \\
		\rowcolor{gray!25} DEA (ours)        & \textbf{50.0} & \textbf{79.9} \\ 
		\bottomrule[1pt]
	\end{tabular}
\end{table}

\begin{figure}[!t]
	\centering
	\includegraphics[width=3.5in]{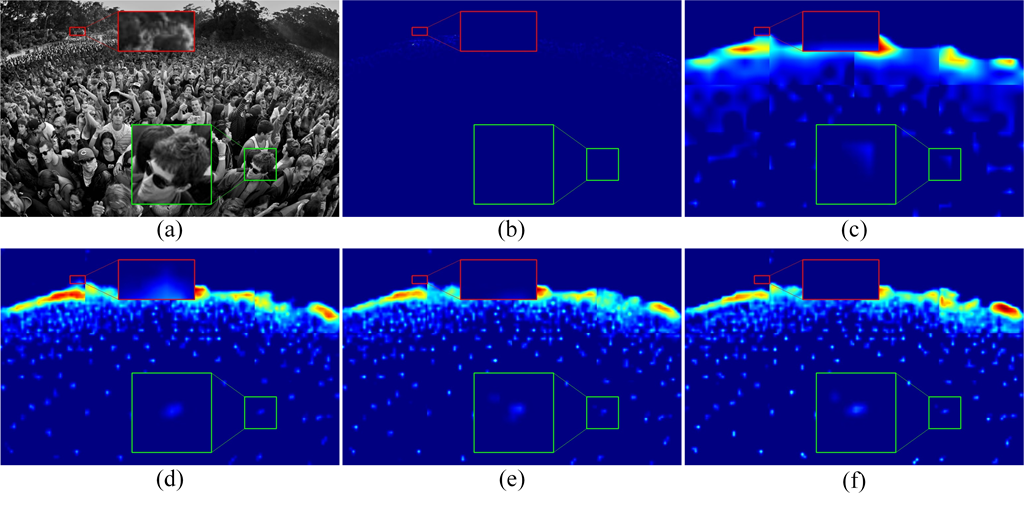}
	\caption{Examples of visualizations of the different modules. (a) Input Image. (b) Ground Truth. (c) Baseline. (d) Baseline + MFFM. (e) Baseline + MFFM + DEAB. (f) Baseline + MFFM+DEAB + ASAM.}
	\label{fig_6}
\end{figure}

{\bf{Effect of Detail Embedded Attention Block.}} We compare the global self-attention (GSA), the global self-attention with integrated local convolutions, and the proposed DEA. As presented in Table \ref{table6}, compared to the global self-attention, DEA showed a 0.7 improvement in MAE and a reduction of 1.8 in MSE. Furthermore, DEA outperforms global self-attention with integrated local convolutions, reducing MAE by 0.8 and MSE by 6.4. These findings underscore the effectiveness of embedding local information into global contexts in a learnable manner, facilitating superior extraction of valuable foreground information from intricate scenes.

\begin{figure}[!t]
	\centering
	\includegraphics[width=3.5in]{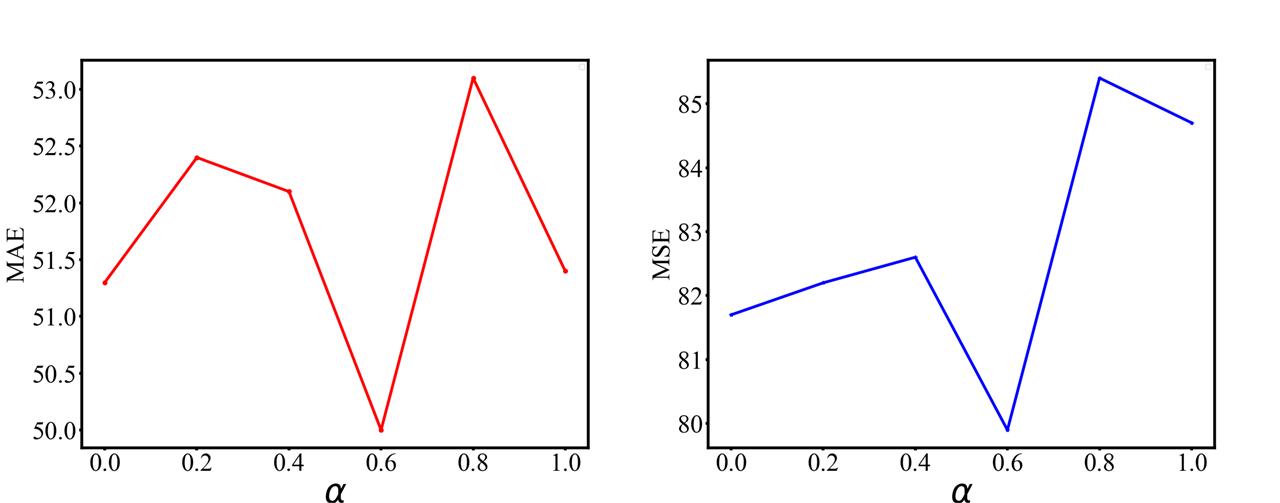}
	\caption{Effect of the value of the learnable parameter $\alpha $.}
	\label{fig_8}
\end{figure}

{\bf{Effect of Learnable Parameter $\bm{\alpha}$.}} We analyze the initialization values of the learnable parameter $\alpha$ in the DEAB. Through comparative experiments, we uniformly sampled six initial values from the range [0, 1] and assessed their effects on counting performance. As depicted in Fig. \ref{fig_8}, initializing $\alpha$ to 0.6 results in the minimum values for both MAE and MSE. Therefore, we set the initialization value of $\alpha$ to 0.6 in this paper.
\begin{table}
	\centering
	\caption{Ablation studies of the different convolution.}
	\renewcommand\arraystretch {1.2}
	\setlength\tabcolsep{5mm}%
	\label{table5}
	\begin{tabular}{lll}
		\toprule[1pt]
		Methods     & MAE $\downarrow$   & MSE $\downarrow$   \\ \hline
		$3\times 3$ Convolution      & 51.5 & \textbf{79.2} \\
		$5\times 5$ Convolution      & \textbf{50.0} & 79.9 \\
		$7\times 7$ Convolution     & 52.7 & 86.7 \\
		\bottomrule[1pt]
	\end{tabular}
\end{table}

{\bf{Effect of Convolutions with Different Sizes.}} We delve into the impact of employing convolutions of different sizes on counting performance within Detail-Embedded Attention. As detailed in Table \ref{table5}, we compare convolutions sized $3\times 3$, $5\times 5$, and $7\times 7$. Interestingly, while the employment of a $5\times 5$ convolution led to a slightly higher MSE compared to the $3\times 3$ convolution, it results in the lowest MAE, outperforming the second-best result by 1.5 points. This highlights the advantage of utilizing a $5\times 5$ convolution for extracting local detail information.

\begin{table}
	\centering
	\caption{Ablation studies of Input-dependent Deformable Conv.}
	\renewcommand\arraystretch {1.2}
	 \setlength\tabcolsep{4mm}%
	\label{table7}
	\begin{tabular}{lll}
		\toprule[1pt]
		Methods     & MAE $\downarrow$  & MSE $\downarrow$  \\ \hline
		Convolution        & 50.8 & 78.0 \\
		Deformable Convolution     & 51.3 & 77.9 \\
		\rowcolor{gray!25} IDConv (ours)   & \textbf{48.3} & \textbf{72.1} \\
		\bottomrule[1pt]
	\end{tabular}
\end{table}

{\bf{Effect of Input-dependent Deformable Convolution.}} In the ASAM, we compare the effects of vanilla convolution, deformable convolution, and our input-dependent deformable convolution, as shown in Table \ref{table7}. Overall, compared to vanilla convolution, IDConv reduces MAE and MSE by 2.5 and 5.9, respectively. MAE decreases by $5.8\%$, and MSE decreases by $7.4\%$ than deformable convolution. This indicates that dynamically changing the convolutional kernel weights and shapes based on the input features is more conducive to improving crowd counting performance.

\section{Conclusion} \label{sec:conclusion}
In this paper, we propose RCCFormer, a novel Transformer-based model designed to address the challenges of scale variation and background interference in crowd counting. Our approach leverages an innovative local-global attention block to extract both local and global contextual information, effectively mitigating the impact of complex backgrounds. Additionally, the adaptive scale-aware module is designed to provide adaptive perception abilities across varying head scales. Extensive experiments conducted on four popular crowd counting datasets, along with a visual analysis on the Shanghai Tech\_A dataset, validate the efficacy of each module and highlight the superior performance of RCCFormer. Notably, our work shows that input-dependent deformable convolution is highly effective for scale-adaptive perception. In future work, we will focus on the challenge of missing labels in crowd counting tasks, particularly concentrating on unsupervised and semi-supervised learning. Our objective is to develop advanced methods capable of automatically extracting features from unlabeled data, thereby ensuring accurate and dependable crowd counting.


\bibliographystyle{unsrt}
\bibliography{reference}


\end{document}